\documentclass[10pt,twocolumn,letterpaper]{article}

\usepackage{cvpr}              

\usepackage{graphicx}
\usepackage{amsmath}
\usepackage{amssymb}
\usepackage{booktabs}
\usepackage{multirow}
\usepackage[symbol]{footmisc}
\usepackage[export]{adjustbox}

\usepackage[pagebackref,breaklinks,colorlinks]{hyperref}

\usepackage[capitalize]{cleveref}
\crefname{section}{Sec.}{Secs.}
\Crefname{section}{Section}{Sections}
\Crefname{table}{Table}{Tables}
\crefname{table}{Tab.}{Tabs.}

\def\cvprPaperID{*****} 
\def\confName{HelloVision}
\def\confYear{2024}

\begin{document}

\title{HelloMeme: Integrating Spatial Knitting Attentions to Embed High-Level and Fidelity-Rich Conditions in Diffusion Models}

\author{Shengkai Zhang\\
{\tt\small songkey@pku.edu.cn}
\and
Nianhong Jiao \\
{\tt\small jnhrhythm@tju.edu.cn}
\and
Tian Li\\
{\tt\small litian215@163.com}
\and
Chaojie Yang\\
{\tt\small 12131243chaojie@gmail.com}
\and
Chenhui Xue \footnotemark[1]\\
{\tt\small great\_xch@shu.edu.cn}
\and
Boya Niu \footnotemark[1] \\
{\tt\small by903033784@163.com}
\and
Jun Gao \\
{\tt\small gaojun55@gmail.com}
\\ \\ {\centering \textbf{HelloGroup Inc.}}
}
\maketitle

\footnotetext[1]{Intern.}

\begin{abstract}
We propose an effective method for inserting adapters into text-to-image foundation models, which enables the execution of complex downstream tasks while preserving the generalization ability of the base model. The core idea of this method is to optimize the attention mechanism related to 2D feature maps, which enhances the performance of the adapter. This approach was validated on the task of meme video generation and achieved significant results. We hope this work can provide insights for post-training tasks of large text-to-image models. Additionally, as this method demonstrates good compatibility with SD1.5 derivative models, it holds certain value for the open-source community. Therefore, we will release the related code (\url{https://songkey.github.io/hellomeme}).
\end{abstract}

\section{Introduction}
\label{sec:intro}

The story begins with our task of generating meme videos, which is similar to video-driven portrait animation methods but comes with several specific requirements. First, the facial expressions and head poses in meme images or videos are often highly exaggerated, adding extra challenges to the task. Second, the technical solution needs to have the potential to extend to half-body or even full-body compositions. Third, the solution must not compromise the generalization ability of the text-to-image foundation model, allowing us to leverage the rich customization methods of the Stable Diffusion \cite{rombach2022highresolutionimagesynthesislatent} base model to enhance the diversity of content generation. To meet these requirements, our solution is as follows:

First, for tasks involving exaggerated facial expression-driven, many existing methods perform well. However, these methods usually require that the head pose in the driving video not be too extreme, as large deviations can easily lead to distortion. In our approach, we separately encode the head pose and facial expressions as 2D feature maps and linear features, and then fuse them using spatial knitting attention mechanisms . The fused features serve as the representation of the driving information. This approach improves performance under conditions of exaggerated expressions and poses.

\begin{figure*}[t]
\begin{center}
\includegraphics[width=1\linewidth]{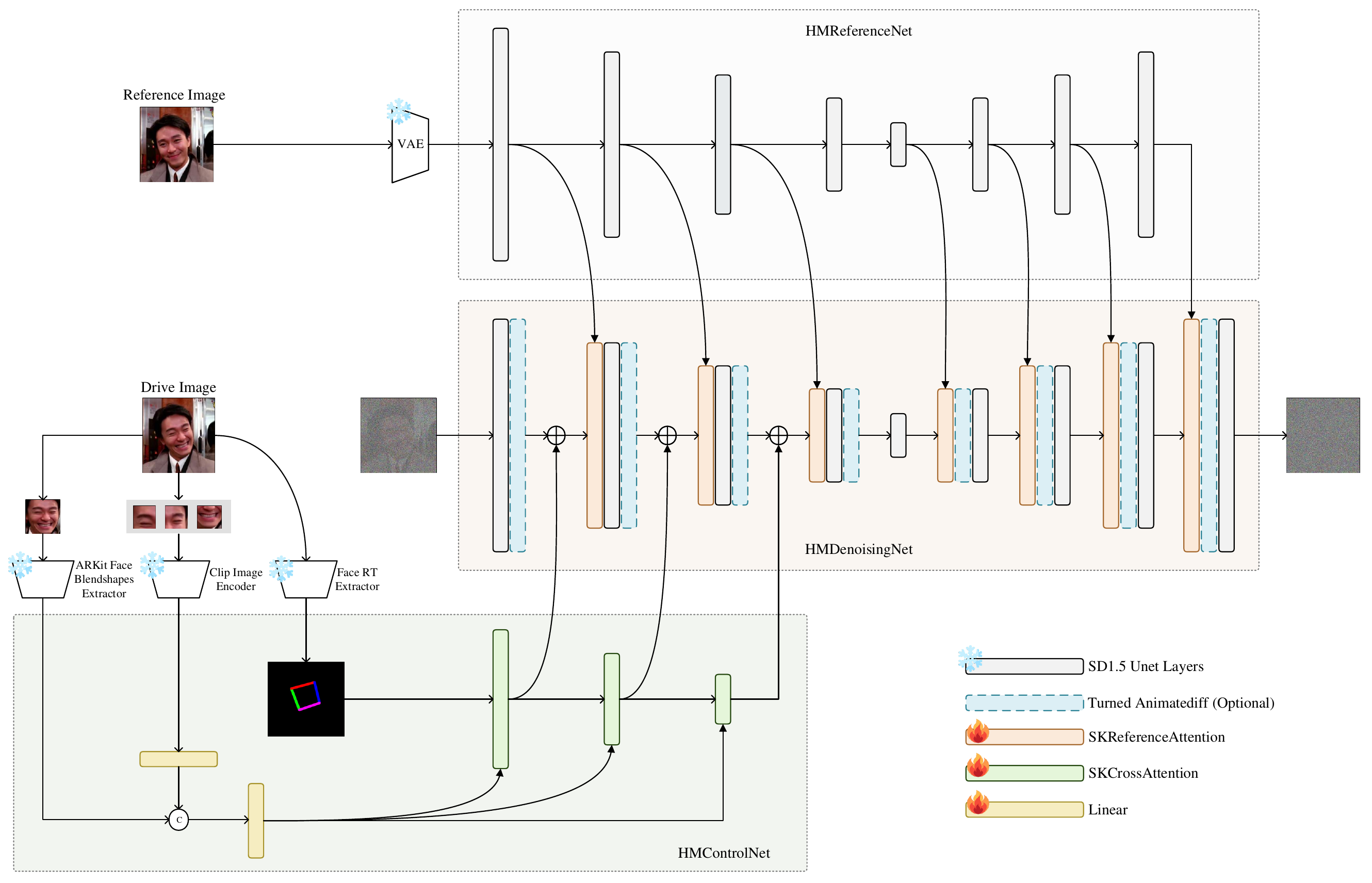}
\end{center}
\caption{Our solution consists of three modules. HMReferenceNet is used to extract Fidelity-Rich features from the reference image, while HMControlNet extracts high-level features such as head pose and facial expression information. HMDenoisingNet receives both sets of features and performs the core denoising function. It can also integrate a fine-tuned Animatediff module to generate continuous video frames.}
\label{fig:structure}
\end{figure*}

Second, for facial reenactment tasks, many works evaluate their performance on body-driven tasks. However, it wasn't until diffusion-based methods \cite{rombach2022highresolutionimagesynthesislatent} emerged that we saw the real promise in addressing both challenges simultaneously. Therefore, we chose to build our solution on the classic SD1.5 model. Based on the results from various Stable Diffusion applications, this technical approach shows ample potential.

Third, Stable Diffusion 1.5 \cite{rombach2022highresolutionimagesynthesislatent} has a significant first-mover advantage, moderate computational requirements, and strong performance, leading to a rich open-source ecosystem—a vast treasure trove of resources. However, we noticed that most current SD-based facial reenactment methods require optimizing all the parameters of the UNet, which compromises compatibility with SD1.5 \cite{rombach2022highresolutionimagesynthesislatent} derived models. Similar to Animatediff \cite{guo2024animatediffanimatepersonalizedtexttoimage} , our approach keeps the SD1.5 UNet weights completely unchanged, optimizing only the inserted adapter's parameters. We found that using a simple adapter struggled to converge, but the introduction of the SK Attention mechanism effectively solved this issue.

\section{Related Work}
\label{sec:rel_work}

\subsection{Condition}

When we use a portrait photo and a set of conditions as input to generate a talking video, this set of conditions can either be a weak condition, such as an audio clip, or a strong condition, such as another talking video. Our task allows the use of the latter, so we focus on discussing the strong condition. If the face bitmap from the talking video is directly used as a condition, it may leak identity information during training. Therefore, the common approach is to extract identity-agnostic driving features from it instead.

a) The most intuitive strong condition is face landmarks, but they couple facial expressions, head pose, and identity information together, and the expression information they represent is often lossy. b) Methods based on 3D face models can theoretically fully decouple facial expressions, head poses, and identity information. However, due to current methods having limitations in the accuracy of 3D coefficient extraction from wild data, the performance of generative solutions that rely on them as an intermediary is inevitably constrained. c) Directly using the aligned bitmap of facial features as a condition is another approach; however, it also contains some identity information, which poses a risk of leakage and needs to be mitigated. d) Methods that use motion latent as a condition are relatively popular because they do not require explicit extraction, thereby avoiding precision limitations. They also allow for end-to-end training, ensuring optimal performance. However, the drawback is that they need to be used in conjunction with GAN-based generators, which limits fidelity under large poses.

Our approach combines the advantages of b) and c), proving to be highly effective in the context of meme video generation.

\begin{figure*}[t]
\begin{center}
\includegraphics[width=1\linewidth]{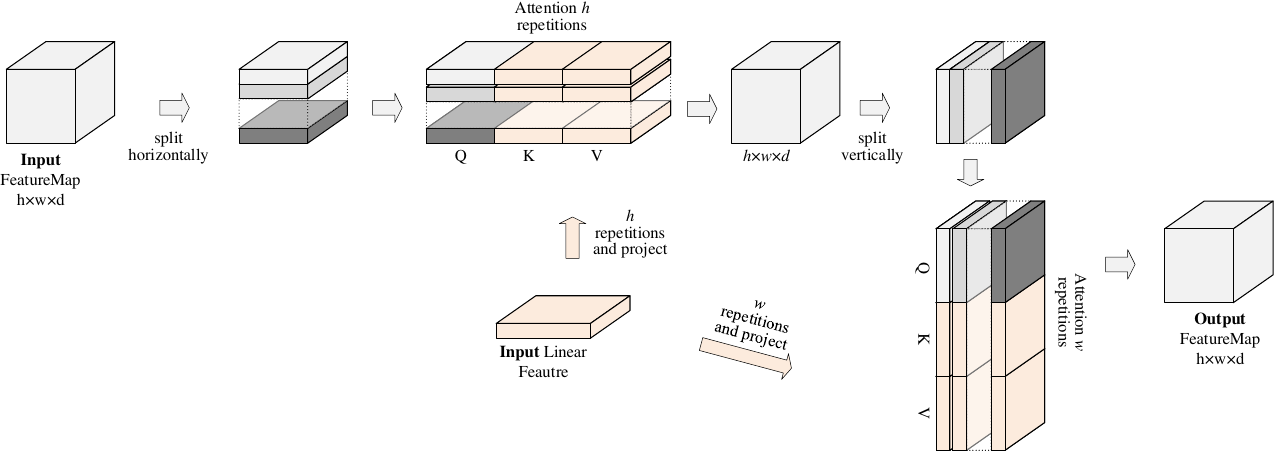}
\vspace{-8mm}
\end{center}
\caption{This is the structural diagram of SKCrossAttention, which utilizes the Spatial Knitting Attention mechanism to fuse 2D feature maps with linear features. It performs cross-attention first row by row, then column by column.}
\label{fig:skca}
\end{figure*}

\subsection{Diffussion/Non-diffusion based Methods}

Currently, the mainstream non-diffusion-based methods are the motion latent-based approaches mentioned earlier, which are characterized by their speed and high fidelity. This allows them to be applied in real-time scenarios that other methods cannot achieve. However, it is essentially a GAN-based method, which lacks sufficient understanding of human body structure. As a result, we can observe that its fidelity in scenarios involving large angles or compositions greater than half-body is less than satisfactory.

Diffusion-based methods are currently advancing rapidly, and we have already seen a wealth of applications that demonstrate the potential of diffusion models to understand the world. Therefore, we need not worry about their insufficient understanding of human structure. The emerging diffusion-based facial reenactment methods have demonstrated high fidelity and generalization ability, with a variety of implementation approaches. I believe this is just the beginning. Currently, diffusion-based methods require significant computational resources and take a considerable amount of time. However, I believe this issue will diminish as algorithms and hardware continue to advance.

\subsection{T2I Base Model Post Training}

Text-to-image post-training methods have made significant progress, enabling increasingly complex downstream tasks. Initially, we observed that methods such as Lora, DreamBooth \cite{ruiz2023dreamboothfinetuningtexttoimage}, Hypernetworks, and Textual Inversion \cite{gal2022imageworthwordpersonalizing} allowed for easy customization of T2I models in specific scenarios. Approaches like ControlNet \cite{zhang2023addingconditionalcontroltexttoimage} and IP-Adapter \cite{ye2023ipadaptertextcompatibleimage} facilitated controllable image generation, while Animatediff \cite{guo2024animatediffanimatepersonalizedtexttoimage} readily extended T2I models to T2V models. Furthermore, benefiting from the preservation of the original UNet capabilities, models trained using these methods can be combined to create various stunning effects.

For more complex tasks such as body-driven animation \cite{hu2024animateanyoneconsistentcontrollable} \cite{xu2023magicanimatetemporallyconsistenthuman} \cite{zhang2024mimicmotionhighqualityhumanmotion}, facial reenactment \cite{rombach2022highresolutionimagesynthesislatent} \cite{wei2024aniportraitaudiodrivensynthesisphotorealistic} \cite{guo2024liveportraitefficientportraitanimation} \cite{xu2024vasa1lifelikeaudiodriventalking}, and virtual try-ons \cite{xu2024tunneltryonexcavatingspatialtemporal} \cite{kim2023stablevitonlearningsemanticcorrespondence} \cite{Gou_2023}, it is common to initialize the weights using base SD UNet, followed by updating all weights during subsequent training. This practice inevitably compromises the generalization ability of the underlying text-to-image model and makes it challenging to maintain compatibility with other SD-derived models.

Our work attempts to use a plugin-based approach for post-training the base T2I model to achieve complex downstream tasks while preserving the generalization ability of the base model.

\section{Method}
\label{sec:method}

As shown in \cref{fig:structure}, our solution consists of three modules: HMReferenceNet, HMControlNet, and HMDenoisingNet. HMReferenceNet is used to extract fidelity-rich features from the reference image and is a complete SD1.5 UNet, which only needs to be executed once during inference. HMControlNet is responsible for extracting high-level features include head poses and facial expressions, which are then mapped to three different scales of the latent space in the UNet. HMDenoisingNet is the core denoising model that based on a complete SD1.5 UNet, receives features from HMReferenceNet and HMControlNet to generate an image that imparts new head poses and facial expressions to the reference image. HMDenoisingNet can also incorporate with a fine-tuned Animatediff \cite{guo2024animatediffanimatepersonalizedtexttoimage}  module to generate videos.

\subsection{Spatial Knitting Attentions}

It can be observed that when performing self-attention on 2D feature maps or cross-attention between 2D feature maps and linear features, the feature map is typically flattened row by row into a linear feature. After the attention operation is completed, it is reshaped back into a 2D feature map. Even though 2D positional encoding can be added after flattening the feature map, this operation still partially disrupts the spatial structure information inherent in the 2D layout.

We modified the operation of directly flattening the 2D feature map for attention by first performing attention row-wise, followed by attention column-wise. In addressing the meme generation task, we found that using the former approach required updating all parameters to achieve slow convergence. However, with the latter approach, updating only the limited parameters of the inserted module was sufficient to achieve good results. The latter process resembles the interweaving of warp and weft threads during weaving, so we refer to this mechanism as Spatial Knitting Attentions (SK Attentions).

We believe the effectiveness of the spatial knitting attentions mechanism lies in its natural preservation of the structural information in the 2D feature map, allowing the neural network to avoid the need to relearn this concept. \cref{fig:skca} and \cref{fig:skrefattrn} illustrate the implementation details of the two SKAttention variants we designed, with \cref{sec:sk_exps} discussing their characteristics and applications.

\subsection{HMReferenceNet}

As previously mentioned, HMReferenceNet is a complete SD1.5 UNet. Inspired by \cite{xu2023magicanimatetemporallyconsistenthuman}, it leverages the SD1.5 UNet's inherent ability to understand visual information while effectively mapping the fidelity-rich features from the reference image to the corresponding hidden layers of HMDenoisingNet. However, in most previous works, the parameters of the ReferenceNet are updated during training, which I believe similarly compromises the capability of the base model. In our approach, all weights of HMReferenceNet are kept fixed. In the code implementation, only minor modifications were needed, so I merged the implementation of HMReferenceNet and HMDenoisingNet in one class.

\subsection{HMControlNet}

Head movements and facial expressions are global and local features at different levels, so they should be encoded and fused using different mechanisms. We extract the rotation and translation (RT) values of the head in camera space to represent head pose. A rectangular box in space is then subjected to RT transformation and perspective transformation, resulting in a 2D rasterized image. The four edges of the rectangle are assigned different colors (as shown in \cref{fig:structure}), and this 2D image can fully represent the information of head movement. As discussed earlier, the accuracy of the RT values has its limits. However, compared to micro-expressions, small errors in head pose are not as perceptually noticeable, making it acceptable.

Although facial expressions are local movements, they convey far more information than head pose, so we use two methods in combination to encode them. First, we trained our own model to extract ARKit facial expression coefficients \cite{ARKit}, capturing 51 coefficients. Its advantage is that it is completely decoupled from identity and captures a fairly complete range of micro-expressions. However, its limitation is the extraction model's accuracy, which has an upper limit, and it tends to make errors under large head poses. To improve accuracy during inference, expression coefficients can also be extracted using ARKit on iOS platforms.

Second, to enhance the expressiveness of missing facial details from the expression coefficients, we encode image patches of the eyes and mouth using a CLIP image encoding module and fuse them with the expression coefficients. As mentioned earlier, the risk here is that the identity of the reference image could potentially leak during training, so we applied strong random blurring to the patch images during training to mitigate this risk.

At this point, the head pose information is encoded into a 2D feature map, and the facial expression information is encoded into linear features. We use SKCrossAttention (\cref{fig:skca}) to fuse them into three scales of feature maps, which are then passed to HMDenoisingNet.

\subsection{HMDenosingNet}

\begin{figure}[t]
\begin{center}
\includegraphics[width=1\linewidth]{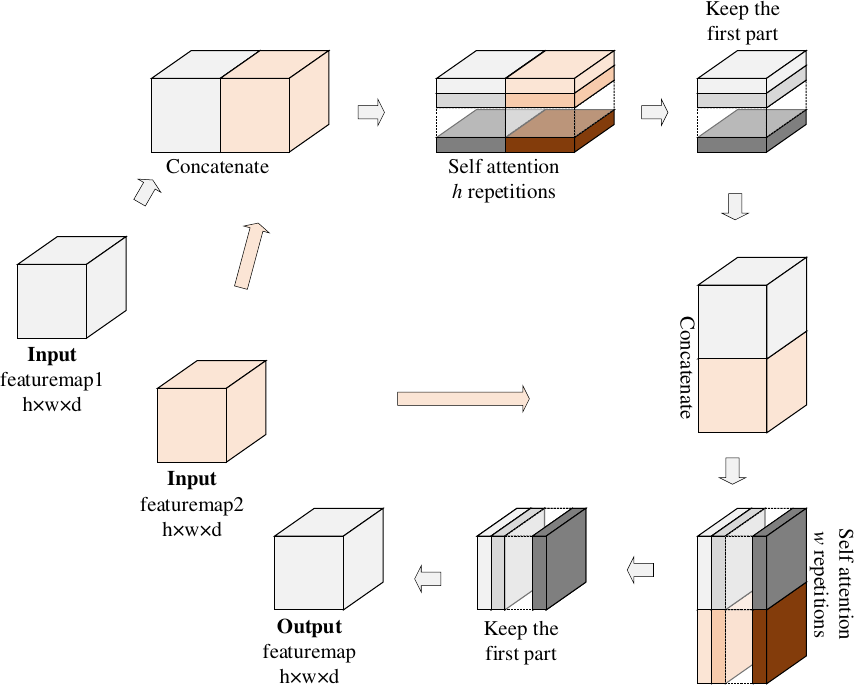}
\vspace{-8mm}
\end{center}
\caption{This is the structural diagram of SKReferenceAttention, which uses the Spatial Knitting Attention mechanism to fuse two 2D feature maps. Specifically, the two feature maps are first concatenated row by row, followed by performing self-attention along the rows. Afterward, only the first half of each row is retained. A similar operation is then performed column by column.}
\label{fig:skrefattrn}
\end{figure}

HMDenoisingNet, built on a complete SD1.5 UNet, receives features passed from HMReferenceNet and HMControlNet. The features from HMReferenceNet are received using SKReferenceAttention (\cref{fig:skrefattrn}), which is inspired by \cite{tian2024emoemoteportraitalive} and incorporates the Spatial Knitting Attention mechanism. The features passed from HMControlNet are directly added to the corresponding feature maps at three scales in the hidden layers of HMDenoisingNet. Both HMControlNet and SKReferenceAttention use the ZeroConvolution mechanism \cite{zhang2023addingconditionalcontroltexttoimage}, ensuring stability during the training process.

Since the weights of the SD 1.5 UNet remain fixed during training, with only the weights of SKReferenceAttention being updated, the model maintains good compatibility with SD 1.5-derived models during inference.

\begin{table*}[t]
\centering
\begin{tabular}{lccccccccc}
\toprule
\multirow{2}{*}{Method} & \multicolumn{5}{c}{Self-Reenactment}  & \multicolumn{4}{c}{Cross-Reenactment} \\
\cline{2-6}  \cline{8-10} 
 & FID$\downarrow$ & FVD $\downarrow$ & PSNR$\uparrow$ & SSIM$\uparrow$ & LPIPS$\uparrow$ & & FID$\downarrow$ & AED$\downarrow$ & APD$\downarrow$ \\ \midrule
Liveportrait\cite{guo2024liveportraitefficientportraitanimation} & 43.84 & 262.19 & 30.66 & 0.649 & 0.228 & & 313.09 & 1.02 & 0.204 \\
Aniportrait\cite{wei2024aniportraitaudiodrivensynthesisphotorealistic} & 38.34 & 384.98 & 30.78 & 0.695 & 0.147 & & 309.52 & 0.96 & 0.068 \\
FollowyourEmoji\cite{ma2024followyouremojifinecontrollableexpressivefreestyle} & 39.11 & 301.71 & 30.91 & 0.695 & 0.152 & & 312.46 & 0.97 & 0.071 \\
\hline
\textbf{Ours} & \textbf{37.69} & \textbf{231.55} & \textbf{31.08} & \textbf{0.704} & \textbf{0.143} & & \textbf{304.35} & \textbf{0.81} & \textbf{0.051} \\
\bottomrule
\end{tabular}
\caption{In comparing our method with the open-source SOTA, it’s important to note that during FVD evaluation, 25 continuous frames are randomly selected from each sample video to calculate the metrics. This leads to variations in the absolute values of test results each time; however, after multiple validations, we found that their relative rankings remain consistent with the values presented in the table.}
\label{tab:exp_results}
\end{table*}

\subsection{Motion}

The components described above enable single image controllable generation. If the driving condition is a continuous video, we can generate frame by frame to achieve controllable video generation. However, using this approach results in significant flickering between video frames. To address the issue of frame discontinuity, we introduced Animatediff's \cite{guo2024animatediffanimatepersonalizedtexttoimage} motion module into HMDenoisingNet to improve inter-frame continuity. However, this reduces the fidelity of the generated video, so we fine-tuned the motion module to enhance the quality.

Animatediff \cite{guo2024animatediffanimatepersonalizedtexttoimage}  generates continuous frames using a 16-frame patch. In our scenario, even with overlapping frames between patches for smoothing, flickering between patches still occurs. Therefore, we divided the video generation process into two stages.

In the first stage, video frames are generated frame by frame with fewer sampling steps, and all frames share the same initial noise. The video generated at this stage has poor continuity and fidelity. In the second stage, the frames generated in the first stage are re-noised and used as the initial noise. Combined with the Motion Module, the frames are generated patch by patch, with overlapping frames used for smoothing between patches. This approach not only allows for the generation of longer videos but also improves both frame continuity and fidelity.

\subsection{Loss}

To enhance the representation of exaggerated facial expressions, we applied weighted loss to the eye and mouth regions. The specific method is similar to the FFG loss described in \cite{ma2024followyouremojifinecontrollableexpressivefreestyle}, and the loss function is expressed as follows:

\begin{equation}
  \mathcal{L}_{LDM} = || z -\hat{z} ||^2
  \label{eq:loss_ldm}
\end{equation}

\begin{equation}
  \mathcal{L} = mean( \mathcal{L}_{LDM}) + sum(\mathcal{M}\cdot\mathcal{L}_{LDM})\cdot \alpha \cdot \beta
  \label{eq:loss}
\end{equation}

Since the details of the eyes and mouth are mostly generated during the perceptual reconstruction stage, we applied greater weight, denoted as $\alpha = (1000-timestep)/1000$, in the later stages of sampling. $\beta = 1 / (sum(\mathcal{M} )+\epsilon)$ is used to normalize the impact caused by varying face sizes in the training data.

\section{Experiments}
\label{sec:exp}

\subsection{Implementations}

\begin{figure*}[htb]
    \centering
    \begin{subfigure}[b]{0.33\textwidth}
        \includegraphics[width=\textwidth]{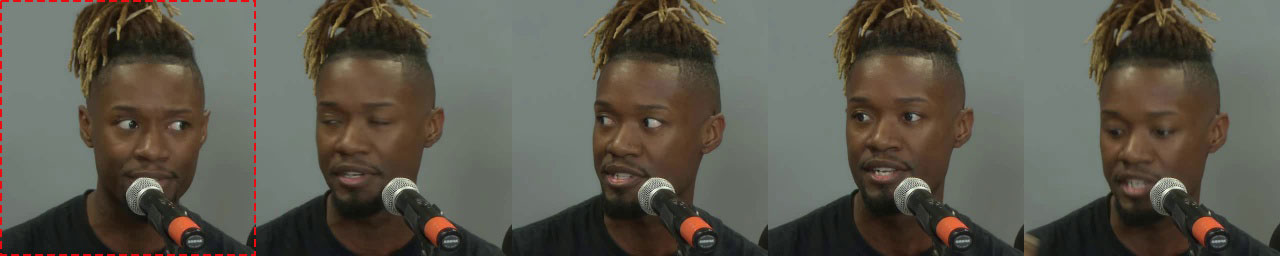}
    \end{subfigure}
    \begin{subfigure}[b]{0.33\textwidth}
        \caption{Ground Truth}
        \includegraphics[width=\textwidth]{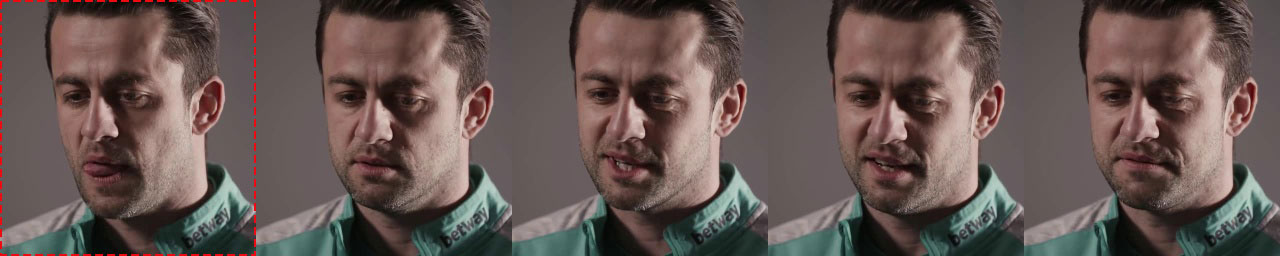}
    \end{subfigure}
    \begin{subfigure}[b]{0.33\textwidth}
        \includegraphics[width=\textwidth]{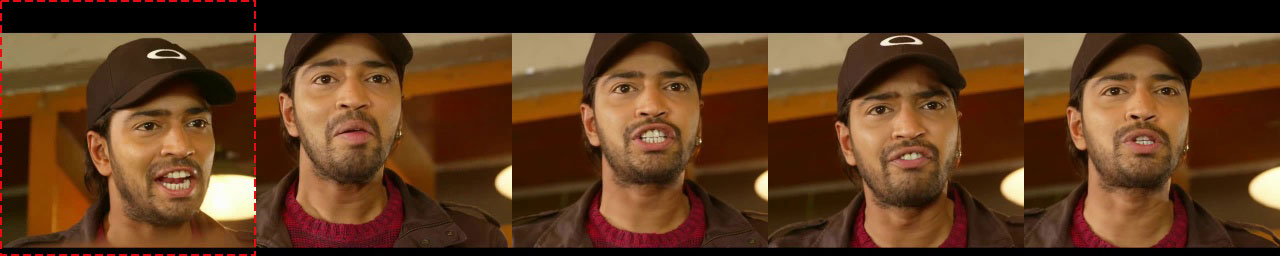}
    \end{subfigure}

    \begin{subfigure}[b]{0.33\textwidth}
        \includegraphics[width=\textwidth]{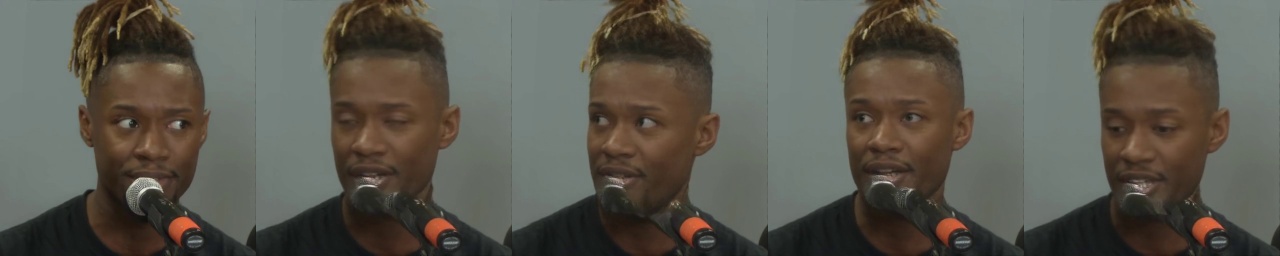}
    \end{subfigure}
    \begin{subfigure}[b]{0.33\textwidth}
        \caption{Liveportrait }
        \includegraphics[width=\textwidth]{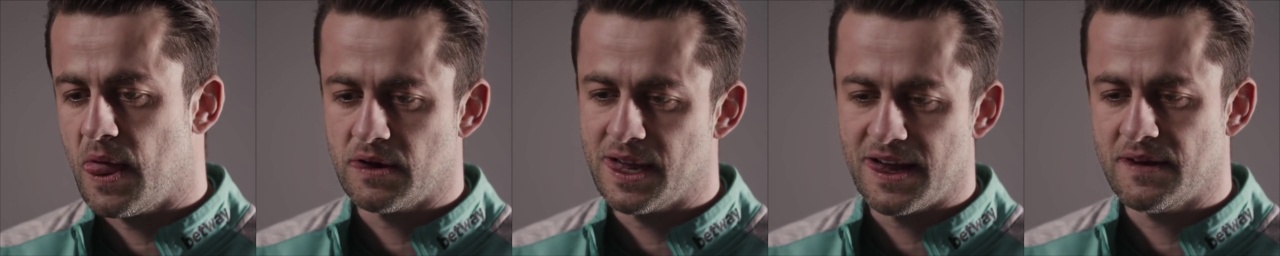}
    \end{subfigure}
    \begin{subfigure}[b]{0.33\textwidth}
        \includegraphics[width=\textwidth]{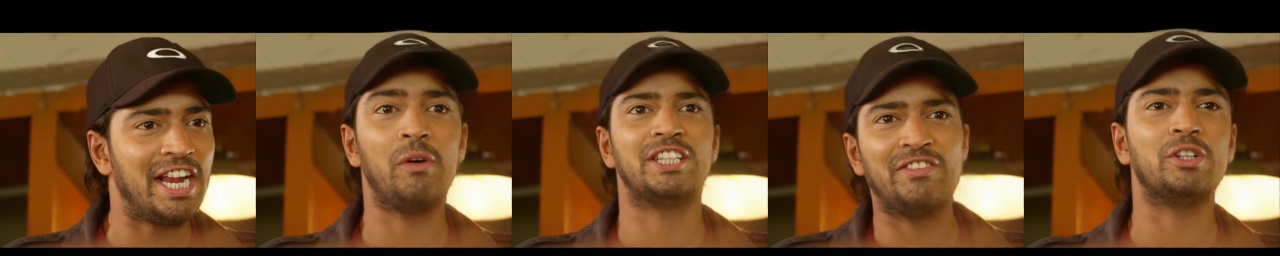}
    \end{subfigure}

    \begin{subfigure}[b]{0.33\textwidth}
        \includegraphics[width=\textwidth]{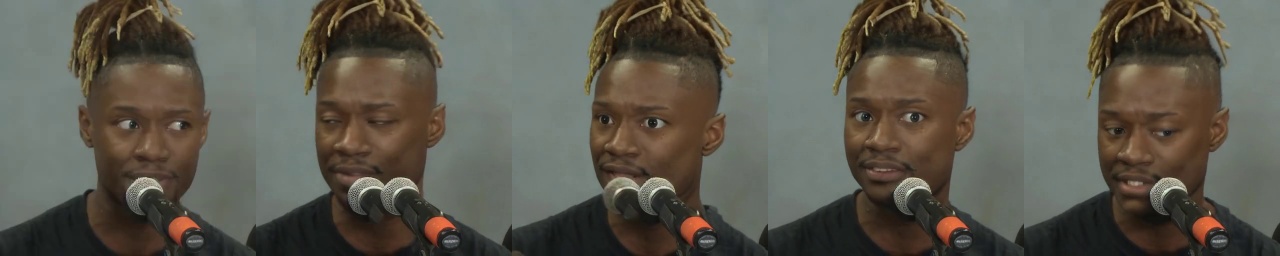}
    \end{subfigure}
    \begin{subfigure}[b]{0.33\textwidth}
        \caption{Aniportrait}
        \includegraphics[width=\textwidth]{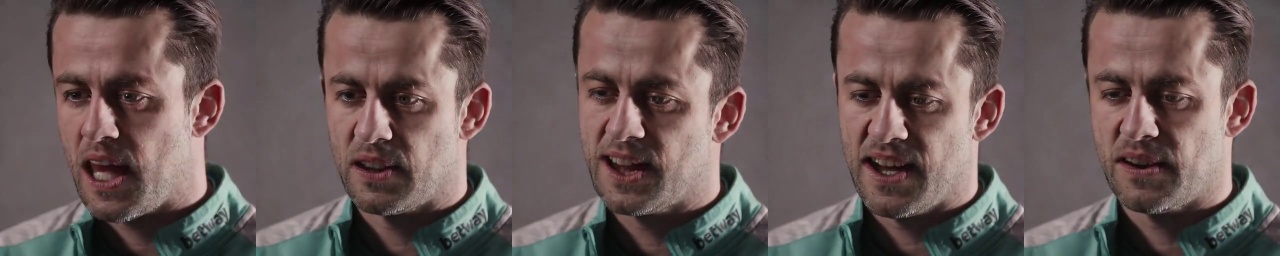}
    \end{subfigure}
    \begin{subfigure}[b]{0.33\textwidth}
        \includegraphics[width=\textwidth]{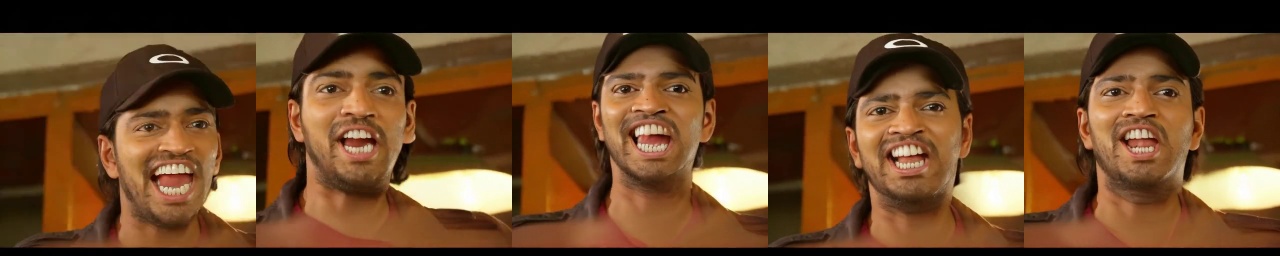}
    \end{subfigure}

    \begin{subfigure}[b]{0.33\textwidth}
        \includegraphics[width=\textwidth]{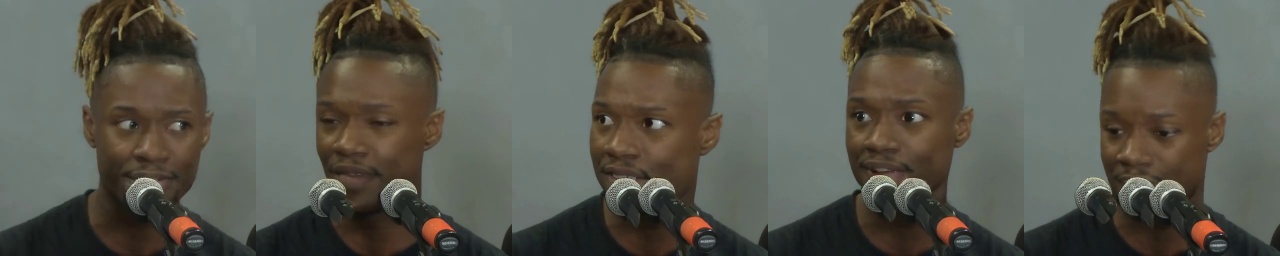}
    \end{subfigure}
    \begin{subfigure}[b]{0.33\textwidth}
        \caption{FollowyourEmoji}
        \includegraphics[width=\textwidth]{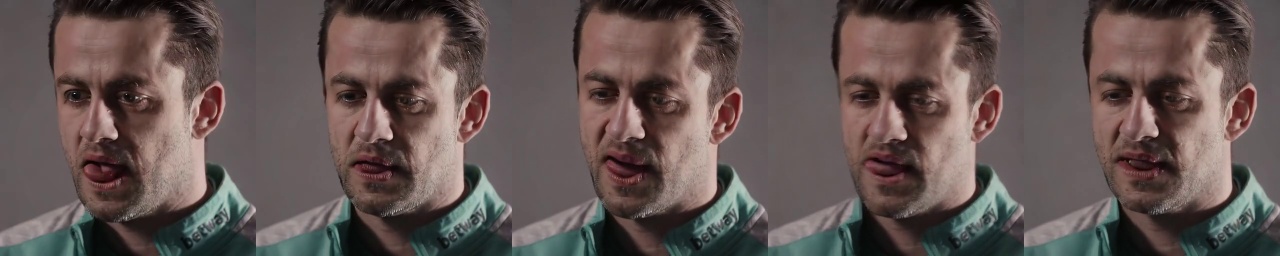}
    \end{subfigure}
    \begin{subfigure}[b]{0.33\textwidth}
        \includegraphics[width=\textwidth]{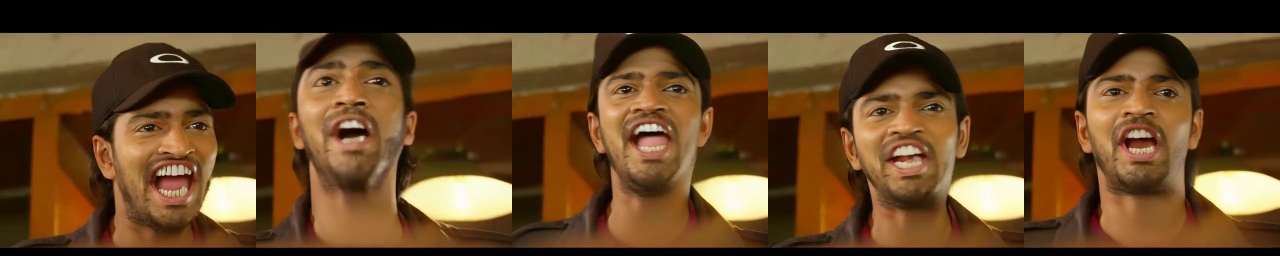}
    \end{subfigure}

    \begin{subfigure}[b]{0.33\textwidth}
        \includegraphics[width=\textwidth]{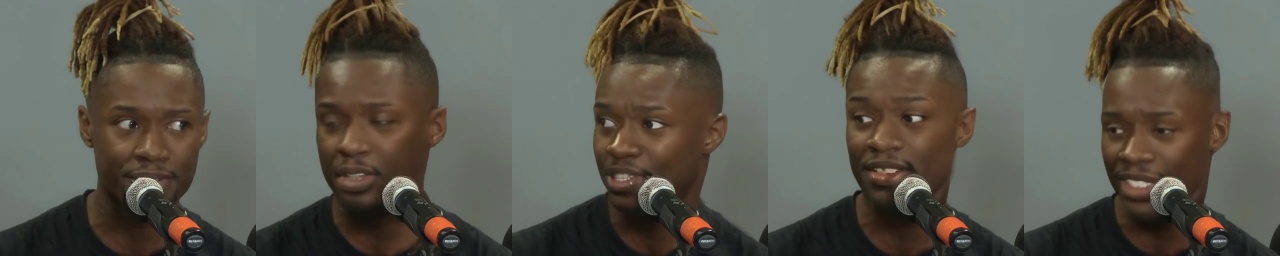}
    \end{subfigure}
    \begin{subfigure}[b]{0.33\textwidth}
        \caption{Ours}
        \includegraphics[width=\textwidth]{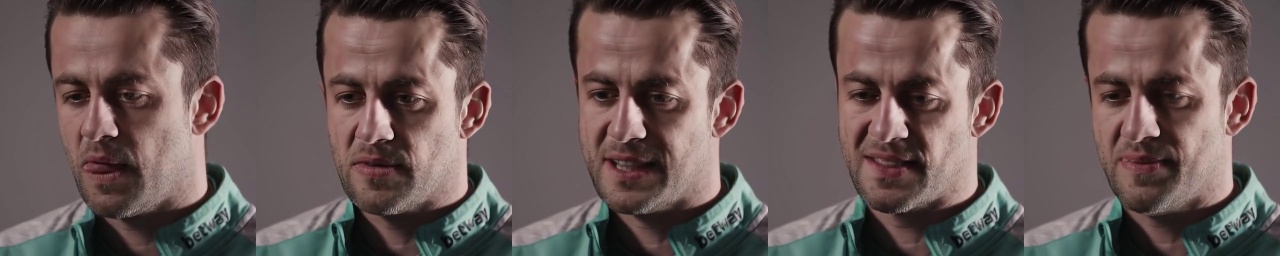}
    \end{subfigure}
    \begin{subfigure}[b]{0.33\textwidth}
        \includegraphics[width=\textwidth]{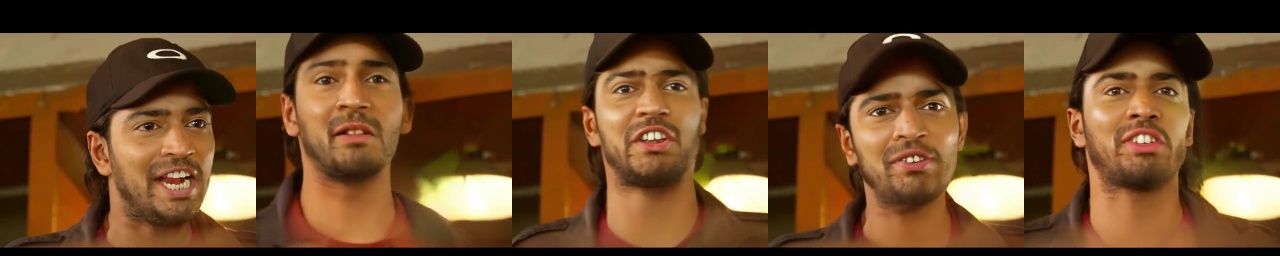}
   \end{subfigure}

    \caption{Examples of self-reenactment performance comparisons, with five frames sampled from each video for illustration. The first row represents the ground truth, with the initial frame serving as the reference image (outlined in red dashed lines).}
    \label{fig:methods_compare}
\end{figure*}

Our training data sources include CelebV-HQ \cite{zhu2022celebvhq}, VFHQ \cite{xie2022vfhq}, and other publicly available videos from the internet. After cropping to a format similar to VFHQ, we manually selected videos with fixed backgrounds. The total dataset amounts to approximately $180$ hours, uniformly preprocessed to a format of $15$ fps and $512 \times 512$ resolution.

During the training of HMControlNet and SKReferenceAttention, a pair of video frames from the same video clip is randomly selected, with one serving as the reference image and the other as the driving image, forming a single training sample. The training process utilized $8$ NVIDIA A100 GPUs, with a batch size of $42$. The learning rate was managed using a Cosine Scheduler, with maximum and minimum values set at $5e-5$ and $1e-7$, respectively. The entire training cycle took approximately one week, totaling $200,000$ iterations.

For the fine-tuning of Animatediff module, $16$ consecutive frames from a single video clip were randomly selected as the driving frame sequence, along with one randomly chosen frame as the reference image, forming a training sample. The training process utilized $8$ NVIDIA A100 GPUs, with an effective batch size of $16$ (with gradient accumulation). The learning rate was managed using a Cosine Scheduler, with maximum and minimum values set at $3e-5$ and $1e-7$, respectively. The training lasted $6$ days, totaling $25,000$ iterations.

Another noteworthy point is that during training, all samples share the same text prompt (as shown below). We aimed for this prompt to be capable of generating a portrait with the original SD1.5 alone, as the reference image already encapsulates ample redundant information.

\textit{(best quality), highly detailed, ultra-detailed, headshot, person, well-placed five sense organs, looking at the viewer, centered composition, sharp focus, realistic skin texture} 

\subsection{Quantitative Comparison}

As shown in \cref{tab:exp_results}, we used two configurations to evaluate the objective metrics for this work. First, we assessed the algorithm's self-reenactment performance using 50 video clips from the VFHQ-Test \cite{xie2022vfhq} dataset, where the first frame of each video served as the reference image, and the entire sequence of frames was used as driving conditions. Second, we evaluated the algorithm's cross-reenactment performance by randomly selecting 20 face images from the FFHQ \cite{karras2019stylebasedgeneratorarchitecturegenerative} dataset as reference images and using the 50 video clips from the VFHQ-Test dataset as driving conditions, resulting in a total of 1,000 generated video outcomes in this setup.

In the two settings, different evaluation metrics were selected. In the self-reenactment setting, each frame has pixel-level ground truth, so we applied metrics such as FID \cite{heusel2018ganstrainedtimescaleupdate}, FVD \cite{unterthiner2019accurategenerativemodelsvideo}, Peak Signal-to-Noise Ratio (PSNR), Structural Similarity Index (SSIM) \cite{wang2004image}, and Learned Perceptual Image Patch Similarity (LPIPS) \cite{zhang2018unreasonableeffectivenessdeepfeatures} to assess video generation quality. In the cross-reenactment setting, since ground truth is unavailable, we used Average Expression Distance (AED) \cite{siarohin2020ordermotionmodelimage} and Average Pose Distance (APD) \cite{siarohin2020ordermotionmodelimage}  to evaluate the consistency of facial expressions and head poses between the driving and generated videos, along with a comparative FID between the driving and generated frames as a supplement.

Based on objective validation metrics, our method consistently outperforms the other three approaches, which aligns with general subjective impressions. It is important to note that the FFHQ dataset has a composition with a high face-to-frame ratio, which differs significantly from the default settings in most face reenactment tasks. As a result, the metrics in the Cross-Reenactment setting are generally somewhat lower than those in the Self-Reenactment setting.

\subsection{Qualitative Comparison}

\cref{fig:methods_compare} presents a comparison of the self-reenactment performance of four methods, selecting three sample groups for illustration. Our method shows superior results in scenarios involving occlusions, complex expressions, and large head movements. Analyzing the results alongside objective metrics, it becomes clear that the expression consistency of FollowYourEmoji and Aniportrait with the driving video is comparatively weaker, while LivePortrait encounters challenges with substantial head movements. However, LivePortrait excels in frame-to-frame smoothness, yielding a seamless quality that objective metrics do not capture and can only be appreciated when comparing continuous video sequences.

\section{Conclusion}
\label{sec:conclusion}

In summary, our proposed method incorporates lightweight plugins into the foundational text-to-image model to enable customization for complex downstream tasks, demonstrating innovation in network structure design. However, several problems still merit further optimization.

Firstly, although we employed a two-stage approach, the frame continuity in the generated videos still lags behind GAN-based solutions. While our module is trained on SD1.5, combining it with SD1.5-derived models customized for portrait generation noticeably improves frame continuity, highlighting a potential direction for improvement. Additionally, post-training based on video generation foundation models may be the most fundamental solution moving forward.

Secondly, the fidelity-rich conditions extracted by HMReferenceNet carry such complete information that when our module is combined with stylized SD1.5-derived models, it significantly diminishes the stylization characteristics. This may be partly due to the fact that our training data primarily features real individuals. Nevertheless, enhancing style expressiveness would make this work even more valuable for applications.

Finally, regarding the driving conditions, we selected a theoretically optimal approach—combining facial expressions with head pose—which has also performed well in practice. However, to prevent ID information leakage during training, we applied a strong random blur to the eye and mouth region bitmaps, which is not a very "natural" solution. Therefore, we believe the current driving conditions are not yet perfect and still have substantial potential for improvement.

{\small
\bibliographystyle{ieee_fullname}
\bibliography{egbib}
}

\clearpage \appendix

\section{Experiments on Spatial Knitting Attentions}
\label{sec:sk_exps}

To investigate the peculiarity of Spatial Knitting Attentions, we conducted three experiments within the limits of available time and resources. One experiment was designed to compare the differences between SKCrossAttention and the default CrossAttention in SD1.5. The other two experiments involved applications of Spatial Knitting Attentions, demonstrating how SKAttentions can implement functionalities similar to IPAdapterFaceID and ControlNet. The relevant code and models will be released at \url{https://github.com/HelloVision/ExperimentsOnSKAttentions}.

\begin{figure}[ht]
    \setlength{\tabcolsep}{0pt}
    \begin{tabular}{cccc}
    \footnotesize\shortstack{SD\_EXP\\ txt2im} \ \  & \includegraphics[width=.28\linewidth,valign=m]{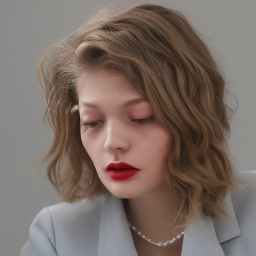} & \includegraphics[width=.28\linewidth,valign=m]{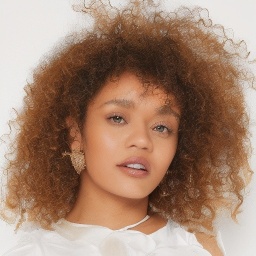} & \includegraphics[width=.28\linewidth,valign=m]{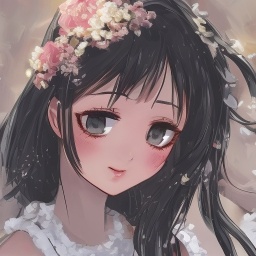}\\[-4pt]
    \footnotesize\shortstack{SK\_EXP\\ txt2im} \ \  & \includegraphics[width=.28\linewidth,valign=m]{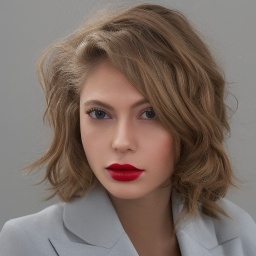} & \includegraphics[width=.28\linewidth,valign=m]{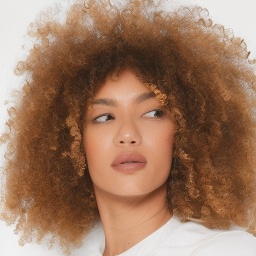} & \includegraphics[width=.28\linewidth,valign=m]{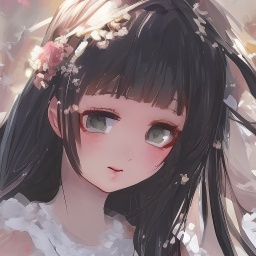}\\
    \footnotesize\shortstack{SD\_EXP\\ im2im} \ \  & \includegraphics[width=.28\linewidth,valign=m]{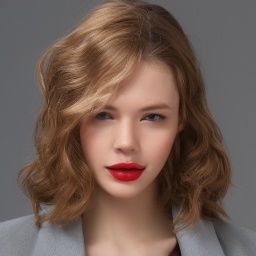} & \includegraphics[width=.28\linewidth,valign=m]{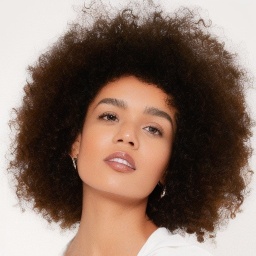} & \includegraphics[width=.28\linewidth,valign=m]{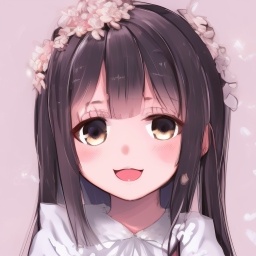}\\
    \footnotesize\shortstack{SK\_EXP\\ im2im} \ \  & \includegraphics[width=.28\linewidth,valign=m]{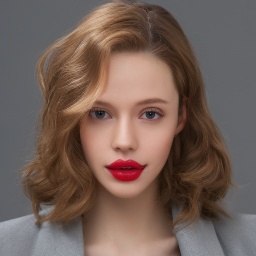} & \includegraphics[width=.28\linewidth,valign=m]{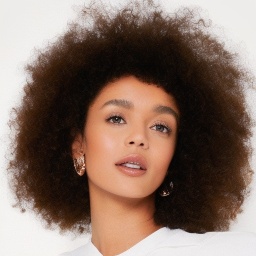} & \includegraphics[width=.28\linewidth,valign=m]{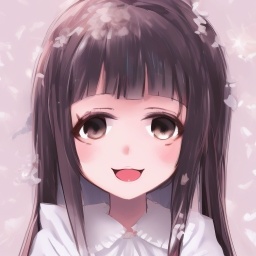}\\
    \footnotesize GT & \includegraphics[width=.28\linewidth,valign=m]{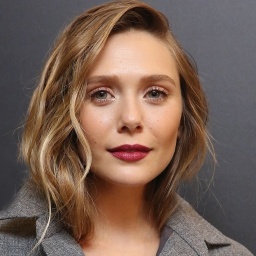} & \includegraphics[width=.28\linewidth,valign=m]{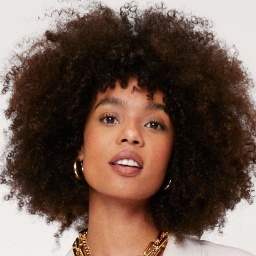} & \includegraphics[width=.28\linewidth,valign=m]{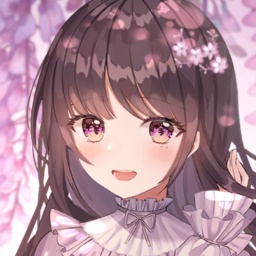}\\
    \end{tabular}
    \caption{SD\_EXP vs. SK\_EXP}
    \label{fig:sd_vs_sk}
\end{figure}

We gathered a portrait-related dataset (MomoFaceTrain1\_390W), containing $3.9$ million images as training data. Each image was captioned with Qwen2-VL-2B \cite{wang2024qwen2vlenhancingvisionlanguagemodels}. Two test sets were used: the first (MomoFaceTest1\_1W) contains $10,000$ images with a similar data distribution as the training set but no overlap. The second (FFHQTest1\_3K) is composed of $3,500$ randomly selected images from the FFHQ dataset.

\begin{figure}[ht]
    \setlength{\tabcolsep}{0pt}
    \begin{tabular}{cccc}
    Condition & \includegraphics[width=.28\linewidth,valign=m]{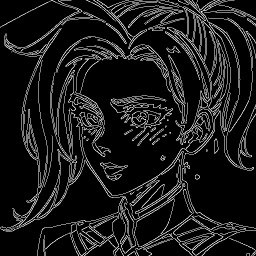} & \includegraphics[width=.28\linewidth,valign=m]{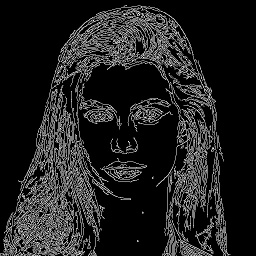} & \includegraphics[width=.28\linewidth,valign=m]{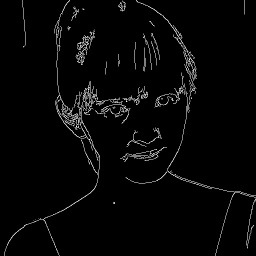}\\
    \footnotesize\shortstack{ControlNet\\ txt2im} \ \  & \includegraphics[width=.28\linewidth,valign=m]{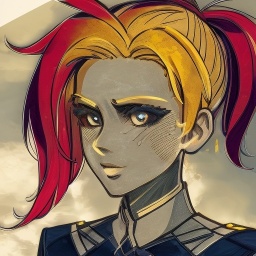} & \includegraphics[width=.28\linewidth,valign=m]{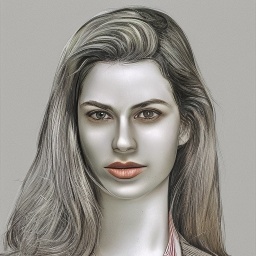} & \includegraphics[width=.28\linewidth,valign=m]{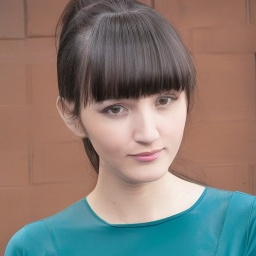}\\
    \footnotesize\shortstack{ControlNet-\\ SK txt2im} \ \  & \includegraphics[width=.28\linewidth,valign=m]{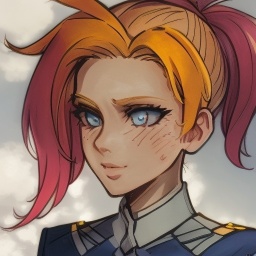} & \includegraphics[width=.28\linewidth,valign=m]{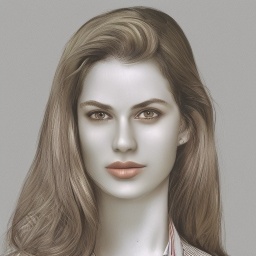} & \includegraphics[width=.28\linewidth,valign=m]{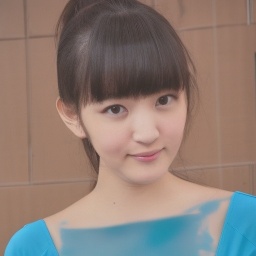}\\
    \footnotesize\shortstack{ControlNet\\ im2im} \ \  & \includegraphics[width=.28\linewidth,valign=m]{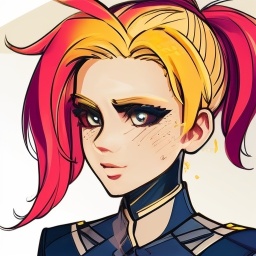} & \includegraphics[width=.28\linewidth,valign=m]{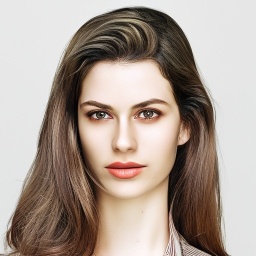} & \includegraphics[width=.28\linewidth,valign=m]{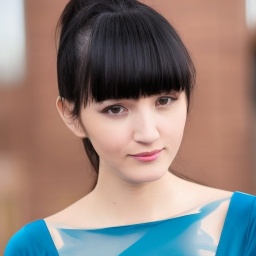}\\
    \footnotesize\shortstack{ControlNet-\\ SK im2im} \ \  & \includegraphics[width=.28\linewidth,valign=m]{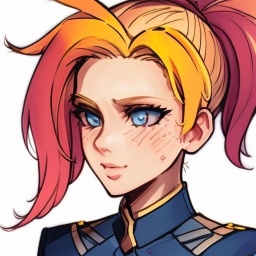} & \includegraphics[width=.28\linewidth,valign=m]{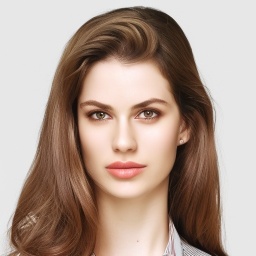} & \includegraphics[width=.28\linewidth,valign=m]{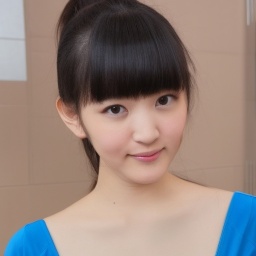}\\
    \footnotesize GT & \includegraphics[width=.28\linewidth,valign=m]{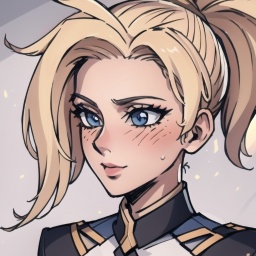} & \includegraphics[width=.28\linewidth,valign=m]{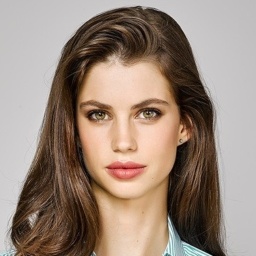} & \includegraphics[width=.28\linewidth,valign=m]{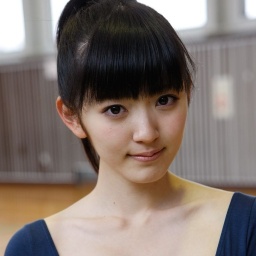}\\
    \end{tabular}
    \caption{ControlNet vs. ControlNetSK}
    \label{fig:cntrl}
\end{figure}

\subsection{Comparison of SKCrossAttention and CrossAttention}

To compare the differences between SKCrossAttention and CrossAttention, we designed two experiments. In the first experiment (SK\_EXP), we replaced CrossAttention in SD1.5 with SKCrossAttention, keeping other network layers and parameters unchanged. The second experiment, serving as a control group (SD\_EXP), randomly reinitialized and trained the weights of CrossAttention in the original SD1.5 model. Both were trained under identical conditions. Specifically, we used $8$ NVIDIA A100 GPUs with batch size of $160$ (with gradient accumulation). The learning rate followed a Cosine Scheduler, with maximum and minimum values of $5e-5$ and $1e-7$, respectively. Training lasted for $42,000$ iterations, with SD\_EXP taking $62$ hours and SK\_EXP taking $72$ hours.

Strictly speaking, it is challenging to ensure complete fairness in this comparison experiment because SK\_EXP requires two CrossAttention operations, resulting in double the number of learnable parameters compared to SD\_EXP. Therefore, achieving fairness is difficult, whether based on the same number of iterations or training time. Another issue is that, due to limited training resources, neither model has fully converged to an optimal state after $42,000$ iterations. Nevertheless, this experiment provides foundational data for understanding the characteristics of Spatial Knitting Attentions, achieving our intended goal.

\begin{figure}[ht]
    \setlength{\tabcolsep}{0pt}
    \begin{tabular}{cccc}
    \footnotesize\shortstack{IP- \\ Adapter\\ txt2im} \ \  & \includegraphics[width=.28\linewidth,valign=m]{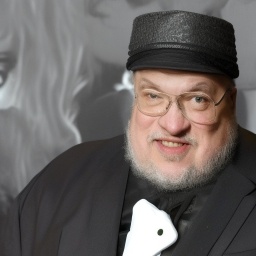} & \includegraphics[width=.28\linewidth,valign=m]{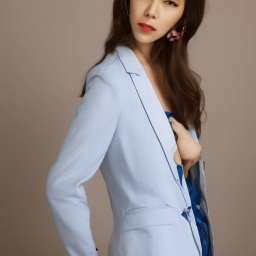} & \includegraphics[width=.28\linewidth,valign=m]{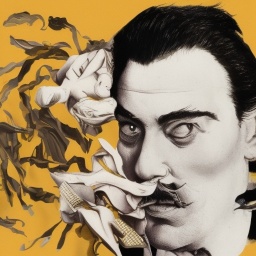}\\
    \footnotesize\shortstack{IP- \\ AdapterSK\\ txt2im} \ \  & \includegraphics[width=.28\linewidth,valign=m]{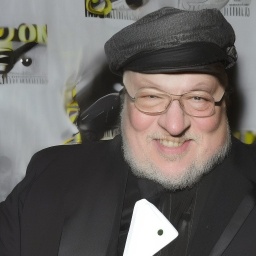} & \includegraphics[width=.28\linewidth,valign=m]{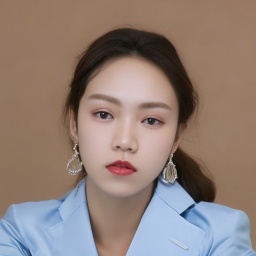} & \includegraphics[width=.28\linewidth,valign=m]{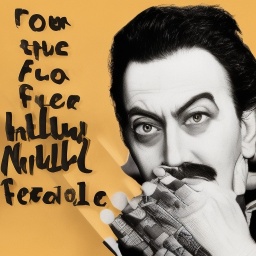}\\[-4pt]
    \footnotesize\shortstack{Mix\\ txt2im} \ \  & \includegraphics[width=.28\linewidth,valign=m]{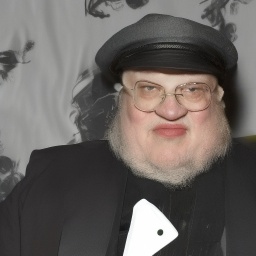} & \includegraphics[width=.28\linewidth,valign=m]{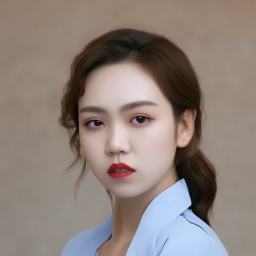} & \includegraphics[width=.28\linewidth,valign=m]{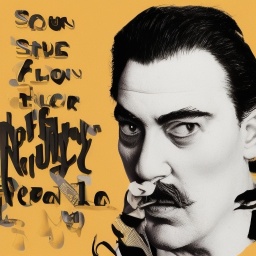}\\
    \footnotesize\shortstack{IP- \\ Adapter\\ im2im} \ \  & \includegraphics[width=.28\linewidth,valign=m]{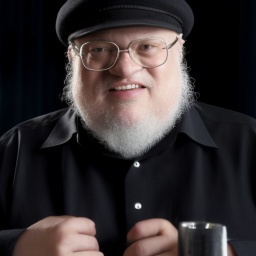} & \includegraphics[width=.28\linewidth,valign=m]{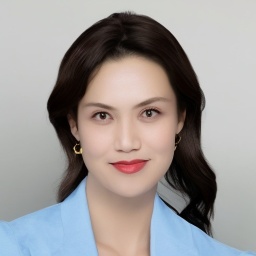} & \includegraphics[width=.28\linewidth,valign=m]{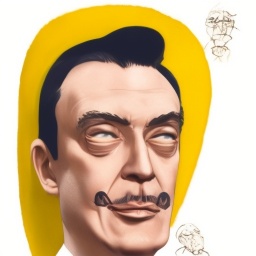}\\
    \footnotesize\shortstack{IP- \\ AdapterSK\\ im2im} \ \  & \includegraphics[width=.28\linewidth,valign=m]{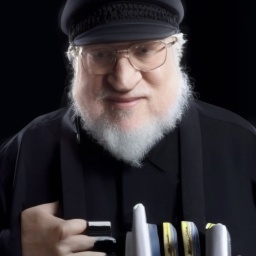} & \includegraphics[width=.28\linewidth,valign=m]{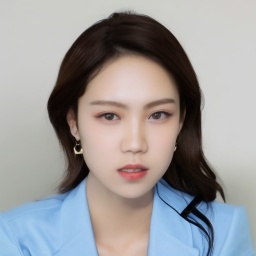} & \includegraphics[width=.28\linewidth,valign=m]{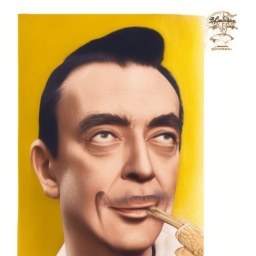}\\[-4pt]
    \footnotesize\shortstack{Mix\\ im2im} \ \  & \includegraphics[width=.28\linewidth,valign=m]{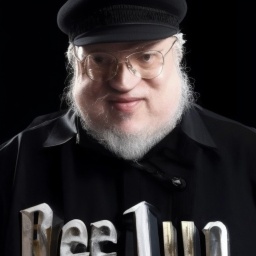} & \includegraphics[width=.28\linewidth,valign=m]{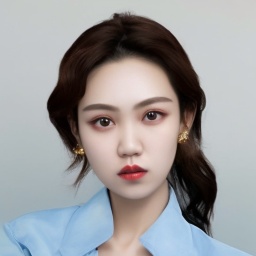} & \includegraphics[width=.28\linewidth,valign=m]{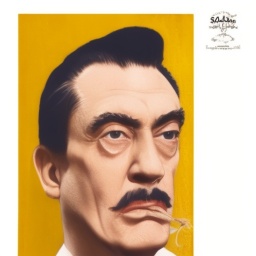}\\
    \footnotesize\shortstack{Reference \\ Image} \ \  & \includegraphics[width=.28\linewidth,valign=m]{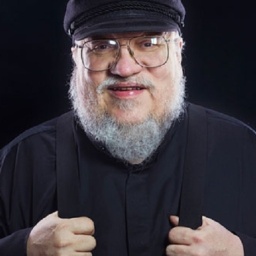} & \includegraphics[width=.28\linewidth,valign=m]{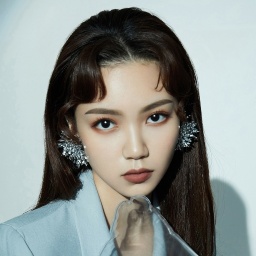} & \includegraphics[width=.28\linewidth,valign=m]{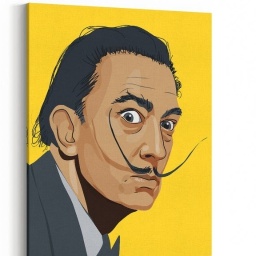}\\
    \end{tabular}
    \caption{IPAdapter vs. IPAdapterSK}
    \label{fig:ipadapter}
\end{figure}

As shown in \cref{tab:sd_vs_sk}, SD\_EXP and SK\_EXP performed comparably across the two test sets, with SK\_EXP slightly outperforming in certain metrics. \cref{fig:sd_vs_sk} displays inference examples from three test cases, and the subjective impressions align well with the objective metrics. We believe that, with sufficient training, the differences between the two methods may become more pronounced. Another possibility is that this experiment essentially reset the original parameters of the base model and retrained it, as discussed in previous experiments on face reenactment tasks. This suggests that SK Attentions might be better suited for use as a plugin. We will continue to verify these hypotheses in future work, and any relevant findings will be updated on the code repository.

\begin{table*}[t]
\centering
\begin{tabular}{lccccccccc}
\toprule
\multirow{2}{*}{Method} & \multicolumn{4}{c}{FFHQTest1\_3K}  & \multicolumn{4}{c}{MomoFaceTest1\_1W} \\
\cline{2-5}  \cline{7-10} 
 & FID$\downarrow$ & PSNR $\uparrow$ & SSIM$\uparrow$ & sim$\uparrow$ & &  FID$\downarrow$ & PSNR$\uparrow$ & SSIM$\uparrow$ & sim$\uparrow$ \\ \midrule
SD\_EXP txt2im & 36.27 & 27.90 & 0.313 & - & & 20.66 & 27.91 & 0.356 & - \\
SK\_EXP txt2im & 35.65 & 27.90 & 0.315 & - & & 19.37 & 27.91 & 0.358 & - \\
SD\_EXP im2im & 28.29 & 28.14 & 0.416 & - & & 13.12 & 28.54 & 0.508 & - \\
SK\_EXP im2im & 28.70 & 28.15 & 0.416 & - & & 12.41 & 28.54 & 0.508 & - \\
\hline
ControlNet txt2im & 24.25 & 27.92 & 0.398 & - & & 21.69 & 27.94 & 0.455 & - \\
ControlNetSK txt2im & 17.99 & 27.91 & 0.471 & - & & 12.39 & 27.91 & 0.548 & - \\
ControlNet im2im & 19.17 & 28.42 & 0.538 & - & & 13.71 & 28.74 & 0.628 & - \\
ControlNetSK im2im & 14.76 & 28.51 & 0.587 & - & & 7.25 & 28.98 & 0.681 & - \\
\hline
IPAdapter txt2im & 68.09 & 27.90 & 0.254 & 0.172 & & 62.22 & 27.91 & 0.285 & 0.154 \\
IPAdapterSK txt2im & 42.08 & 27.89 & 0.290 & 0.195 & & 25.13 & 27.91 & 0.346 & 0.338 \\
Mix txt2im & 38.95 & 27.89 & 0.291 & 0.372 & & 25.96 & 27.92 & 0.344 & 0.440 \\
IPAdapter im2im & 27.51 & 28.13 & 0.391 & 0.262 & & 20.51 & 28.44 & 0.486 & 0.240 \\
IPAdapterSK im2im & 30.75 & 28.13 & 0.399 & 0.213 & & 15.10 & 28.54 & 0.511 & 0.378 \\
Mix im2im & 29.18 & 28.14 & 0.401 & 0.399 & & 15.45 & 28.53 & 0.508 & 0.479 \\
\bottomrule
\end{tabular}
\caption{Evaluation results for the SKAttentions-related experiments, where the "sim" metric represents the similarity between the faces in the reference image and the generated output.}
\label{tab:sd_vs_sk}
\end{table*}

\subsection{Application of SKAttentions}

We found that the SKReferenceAttention module, as described earlier, can easily replicate functions similar to those of ControlNet \cite{zhang2023addingconditionalcontroltexttoimage}, and SKCrossAttention module can similarly emulate the features of IPAdapterFaceID \cite{ye2023ipadaptertextcompatibleimage}. Consequently, we conducted two additional experiments (ControlNetSK and IPAdapterSKFaceID) to validate the broader applicability of SKAttentions. However, due to differences in data distribution, network structure, and training conditions, a fair comparison with the official ControlNet and IPAdapterFaceID models is hard. Thus, these experiments and corresponding comparisons should be considered as rough references only.

\subsubsection{ControlNetSK}

Referring to \cref{fig:structure}, the implementation of ControlNetSK can directly utilize the HMReferenceNet and HMDenosingNet modules. The conditioning image for ControlNet can be extracted directly using the SD1.5 UNet, with only the weights of SKReferenceAttention updated during training. This approach does not require updating the parameters of the entire downsampling and intermediate modules in the UNet as ControlNet.

In the experimental process, we used the Canny edge conditioning from ControlNet to validate feasibility. The training was conducted on $8$ NVIDIA A100 GPUs with an effective batch size of $200$ (using gradient accumulation). The learning rate followed a Cosine Scheduler, with a maximum value of $5e-5$ and a minimum of $1e-7$. The training totaled $42,000$ iterations and took $139$ hours.

As shown in the results in \cref{tab:sd_vs_sk}, ControlNetSK outperforms ControlNet, which is also evident from the visual examples in \cref{fig:cntrl}. This improvement is due to ControlNetSK’s training on portrait data, whereas ControlNet is a more general model, so it’s expected that ControlNetSK would perform better in portrait generation scenarios. Moreover, ControlNetSK achieves these results with fewer learnable parameters and training steps, further validating the effectiveness of this structure.

\subsubsection{IPAdapterSK FaceID}

Similar to IPAdapter FaceID, we used face features extracted by InsightFace \cite{an_2022_pfc_cvpr} to represent face ID. These features are decoupled from attributes like lighting, artistic style, and pose, allowing for effective integration with text prompts and offering significant creative freedom. However, unlike IPAdapterFaceID, we directly replicated the face features five times to form a linear feature of length five, then incorporated face ID information into the UNet using the SKCrossAttention mechanism.

The training process and conditions were similar to those of ControlNetSK. We used $8$ NVIDIA A100 GPUs, with an effective batch size of $224$ (using gradient accumulation). The learning rate followed a Cosine Scheduler, with a maximum value of $5e-5$ and a minimum of $1e-7$. A total of 42,000 iterations were conducted, taking approximately $102$ hours.

As seen from the results in \cref{tab:sd_vs_sk} and the examples in \cref{fig:ipadapter}, IPAdapter and IPAdapterSK perform at similar, relatively moderate levels. However, when used in combination, they produce a substantial improvement in results. This suggests that combining ad adapters for text-to-image base models does not necessarily lead to mutual interference; instead, it can result in mutual enhancement. Furthermore, IPAdapterSK achieving this performance with limited training suggests significant untapped potential for further development.

\subsection{Conclusion}

Our experiments validated the characteristics and potential applications of SKAttentions, demonstrating the value of this structure to some extent. However, there are two areas for improvement. Firstly, the training dataset contains approximately one-third low-resolution data, and we plan to continuously enhance both the quantity and quality of training data. Secondly, insufficient training is another limitation; in the future, we will keep iterating on the related models, and improved results will be updated on the code page. We also intend to explore new applications of this structure, such as other versions of ControlNetSK under different conditions.

\end{document}